\documentclass[conference]{IEEEtran}
\IEEEoverridecommandlockouts
\usepackage{cite}
\usepackage{amsmath,amssymb,amsfonts}
\usepackage{algorithmic}
\usepackage{graphicx}
\usepackage{textcomp}
\usepackage{xcolor}
\usepackage{booktabs}
\usepackage[most]{tcolorbox}

\newif\ifcomments
\commentstrue 
\ifcomments
  \usepackage{pdfcomment}
  \newcommand{\stickynote}[1]{\pdfcomment[icon=Note,color={1 1 0}]{#1}}
\else
  \newcommand{\stickynote}[1]{}
\fi
\def\BibTeX{{\rm B\kern-.05em{\sc i\kern-.025em b}\kern-.08em
    T\kern-.1667em\lower.7ex\hbox{E}\kern-.125emX}}

\begin{document}

\title{Serving Masked Diffusion LLMs: Characterization and Design Principles from Real Hardware}

\author{
\IEEEauthorblockN{Farhana Amin}
\IEEEauthorblockA{\textit{Dept. of Computer Science} \\
\textit{Virginia Tech}\\
Blacksburg, VA, USA \\
afarhana@vt.edu}
\and
\IEEEauthorblockN{Sabiha Afroz}
\IEEEauthorblockA{\textit{Dept. of Computer Science} \\
\textit{Virginia Tech}\\
Blacksburg, VA, USA \\
sabihaafroz@vt.edu}
\and
\IEEEauthorblockN{Mona Moghadampanah}
\IEEEauthorblockA{\textit{Dept. of Computer Science} \\
\textit{Virginia Tech}\\
Blacksburg, VA, USA \\
monamp@vt.edu}
\and
\IEEEauthorblockN{Dimitrios S. Nikolopoulos}
\IEEEauthorblockA{\textit{Dept. of Computer Science} \\
\textit{Virginia Tech}\\
Blacksburg, VA, USA \\
dsn@vt.edu}
}

\maketitle

\begin{abstract}
Masked diffusion language models (dLLMs) can in principle generate text faster than autoregressive (AR)
models, since they denoise many tokens at once. Recent systems have begun building serving infrastructure
for dLLMs, but none first measure how these models behave under real, concurrent serving load. Serving
systems built without this grounding risk carrying over assumptions from AR serving that may not hold for
dLLMs. We characterize dLLM serving to close this gap, using LLaDA-8B-Instruct with a D2F (Discrete
Diffusion Forcing) LoRA adapter on a single NVIDIA H200 GPU, evaluated on GSM8K (tier structure,
predictability, batching, scheduling) and HumanEval (block-size invariance and variance structure). We
report three findings. First, request difficulty, the number of denoising steps a request needs, is
discrete rather than continuous: requests fall into 11 fixed step-count levels ($178 + 29k$), and no signal
we test predicts the level before generation starts (best $R^2 = 0.150$, far below the $0.5$ floor we treat
as scheduling-useful). Second, benchmarks with short generation budgets (\textless\,320 tokens) understate
serving variance, since requests are cut off before the latency spread appears. Third, only 24\% of
single-request wall-clock time is GPU computation; the rest is CPU-side dispatch overhead. Batching mainly
helps by amortizing this overhead: sharing one forward pass per denoising step improves throughput by
16.0$\times$ at batch size 16 over a per-request-dispatch baseline. We also argue structurally that output
quality should not degrade with batch size, stating three assumptions this rests on; we measure 74--76\% GSM8K accuracy at single-request scale. Finally, we derive a
batch-timeout rule for fixed-fill synchronized batching under Poisson arrivals. Together, these results
show that serving diffusion language models needs parallelism at the level of each denoising step, which
differs from AR serving in how admission and eviction interact with an already-shared forward pass, not in
whether a forward pass is shared at all.
\end{abstract}


\section{Introduction}
Masked diffusion language models (dLLMs)~\cite{austin2021structured,lou2024discrete} generate text in a
different way from standard autoregressive (AR) models~\cite{vaswani2017attention}. Instead of writing
tokens one by one from left to right, they start from a fully masked sequence and slowly denoise it over
several steps. Because one forward pass can update a whole block of tokens at once, dLLMs promise higher
throughput than AR decoding. Recent work such as D2F (Discrete Diffusion Forcing)~\cite{wang2025diffusion}
delivers part of this promise, combining block-wise KV caching with adaptive scheduling to reach large
speedups. However, these gains have only been shown for a single request running alone.

The shift toward LLM inference as shared infrastructure is already underway at HPC facilities. Department of Energy (DOE) and
academic centers increasingly run LLM inference as an allocation-based service rather than a commercial
cloud offering. The Argonne Leadership Computing Facility's (ALCF) Inference Service, for example, gives
researchers across the DOE laboratory ecosystem cloud-like access to LLMs running directly on HPC
systems~\cite{alcf_inference_2026}. The workload
properties we study in this paper connect directly to what these facility operators care about. For dLLMs
specifically, per-request compute cost cannot be predicted at admission time (\S\ref{sec:tiers}): unlike AR
inference, where cost scales roughly with prompt and output length known or estimable in advance, a dLLM
request's denoising-step count is not observable until generation is underway. This unpredictability makes
backfill scheduling~\cite{mualem2001backfilling}, which depends on reasonably accurate runtime estimates to
pack jobs into scheduling gaps, harder to apply to dLLM workloads, and it complicates service-level
agreement (SLA) commitments on shared, allocation-managed hardware. The latency-optimal operating point we find at $\rho \approx 0.70$
(\S\ref{sec:scheduling}) gives operators a concrete utilization target for running dLLM endpoints. We
therefore treat this work as a characterization of dLLM serving as a facility workload, not as a
cloud-serving benchmark.

Most existing work on speeding up dLLMs, however, only looks at this single-request case. Methods such as
dKV-Cache \cite{ma2025dkvcache}, Fast-dLLM \cite{wu2025fastdllm}, dLLM-Cache \cite{liu2025dllmcache},
FlashDLM \cite{hu2025flashdlm}, Sparse-dLLM \cite{song2025sparsedllm}, and Spiffy \cite{agrawal2025spiffy}
all reuse computation across denoising steps, through caching, sparsity, or speculation, and each reports
large reductions in per-request latency. What remains open is how these models behave under real serving
conditions, where many requests share one GPU at the same time and each needs a different number of
denoising steps.

Modern AR serving systems, such as vLLM~\cite{kwon2023efficient} and Orca~\cite{yu2022orca}, rely on one key
property of AR decoding: each request moves forward by one token per step, mostly on its own, without
depending on the other requests in its batch. This independence is exactly what makes continuous batching
simple and effective for AR systems. We show that masked diffusion serving does not have this property.
Applying AR-style continuous batching to it without change can make throughput worse, not better.

Here is why. If a continuous batcher were applied directly to masked diffusion decoding, each active request
would still need its own forward pass at every denoising step, because dLLM requests do not share the same
simple, uniform update that AR next-token decoding does. Requests also need different numbers of denoising
steps (\S\ref{sec:tiers}). Together, this means more GPU dispatch calls per step for every active request,
instead of the single shared forward pass that masked diffusion serving actually allows
(\S\ref{sec:design}).

We use two terms in this paper, and define them here so they are clear before \S\ref{sec:design}.
\emph{Continuous batching} is the standard AR-serving approach used by vLLM~\cite{kwon2023efficient} and
Orca~\cite{yu2022orca}: active requests share one forward pass per iteration, and finished requests are
swapped out for new ones as they arrive. \emph{Synchronized batching} is different: all requests in a batch
move forward together, through one shared forward pass per denoising step. This is not a serving strategy
we propose; it is the natural consequence of how block-parallel masked-diffusion decoding already computes
each step (\S\ref{sec:background}), and is the mode existing dLLM inference code, including the D2F
implementation we build on, already executes when a batch is dispatched together. We adopt the term to name
it precisely and contrast it with continuous batching, not to claim priority over the mechanism itself.

A small number of recent systems have started to address dLLM serving directly. dInfer~\cite{ma2025dinfer}
gives a modular inference framework for batched throughput. DiLaServe~\cite{dilaserve2026},
Sangam~\cite{kedia2026sangam}, and dLLM-Serve~\cite{fan2025memoryfootprint} look at scheduling,
prefill/decode partitioning, and memory management, respectively. Our goal is different and
complementary. Rather than build a serving system, we give an empirical, measurement-grounded
characterization of \emph{why} dLLM serving behaves the way it does. Concretely, this characterization
establishes four things not previously reported: that dLLM difficulty is discrete rather than continuous
and cannot be predicted at admission time (\S\ref{sec:characterization}); that per-request dispatch overhead
tracks batch size closely, within 20 to 27\% at smaller batch sizes and matching almost exactly by $B=8$
(\S\ref{sec:design}); that 76\% of single-request latency is host-side dispatch rather than device compute
(\S\ref{sec:profiling}); and that short generation-length benchmarks systematically understate real serving
variance (\S\ref{sec:characterization}). These are mechanistic explanations,we use
them to derive concrete, testable principles, the batch-timeout stability rule (\S\ref{sec:scheduling}) and
the step-level-parallelism argument against naive continuous batching for this workload class
(\S\ref{sec:design}), that existing and future dLLM serving systems, including those cited above, could use
as measurement-grounded design inputs rather than assumptions carried over from AR serving.

Our measurements come from a single GPU. Two
categories of finding generalize differently to shared, multi-tenant HPC facilities.

First, the workload-level properties we characterize, discrete, admission-time-unpredictable difficulty
tiers (denoising step counts cluster into 11 discrete values that cannot be inferred before generation
begins, \S\ref{sec:characterization}), high service-time variance, and the step-level dispatch cost
structure (\S\ref{sec:design}), are properties of the model's computation itself, not of the single-GPU
measurement setup. A cluster scheduler allocating GPUs to dLLM jobs faces the same unpredictable,
high-variance, discretely-tiered service times regardless of whether that GPU is dedicated or shared; these
properties are inputs a multi-tenant scheduler would need to account for, and single-GPU measurement is the
correct way to isolate them from confounding cluster-level effects.

Second, effects that arise specifically from sharing, contention for memory bandwidth or streaming
multiprocessor (SM) resources between co-located jobs, scheduler preemption, and multi-GPU parallelism
strategies for serving larger models, are not measured here and do not follow directly from our results. We propose this characterization as future work.

The batch-timeout stability rule we derive (\S\ref{sec:scheduling}) should be read in this light: it is a
single-GPU, single-tenant service model, not a complete multi-tenant scheduling policy. Its role for
facilities running cluster schedulers such as Slurm~\cite{yoo2003slurm} is as a building block, a per-job
service-time characterization that a multi-tenant scheduler could take as input when making admission and
allocation decisions, rather than a drop-in replacement for existing cluster scheduling logic. A scheduler
that instead assumes AR-style, predictable per-token service costs would misallocate GPU-hours for dLLM
jobs regardless of tenancy model, which is the more general point our single-GPU results support;
validating the stability rule itself under real multi-tenant contention is the natural next step.

Concretely, we make three measurement-based observations, and we use them to motivate a serving design that
departs from AR practice. Each finding pairs a measured result with the mechanism that produces it, and the
design implication it motivates:

\begin{tcolorbox}[
  colback=gray!5, colframe=gray!60, boxrule=0.5pt, arc=1pt,
  left=6pt, right=6pt, top=4pt, bottom=4pt
]
\textbf{1. dLLM request difficulty is discrete, not continuous} (\S\ref{sec:tiers}) \\
\textit{Mechanism:} block-addition rule, new blocks start only at fixed completion checkpoints. \\
\textit{Implication:} admission-time signals cannot predict difficulty; scheduling must treat cost as
unpredictable per-request.
\end{tcolorbox}

\begin{tcolorbox}[
  colback=gray!5, colframe=gray!60, boxrule=0.5pt, arc=1pt,
  left=6pt, right=6pt, top=4pt, bottom=4pt
]
\textbf{2. Short budgets understate variance} (\S\ref{sec:truncation}) \\
\textit{Mechanism:} truncation before natural completion length hides the real latency spread. \\
\textit{Implication:} benchmarks must run to natural completion length ($\sim$320 tokens here) to report
representative variance.
\end{tcolorbox}

\begin{tcolorbox}[
  colback=gray!5, colframe=gray!60, boxrule=0.5pt, arc=1pt,
  left=6pt, right=6pt, top=4pt, bottom=4pt
]
\textbf{3. CPU dispatch, not GPU compute, dominates latency} (\S\ref{sec:profiling}) \\
\textit{Mechanism:} the per-block control loop (masking, thresholds, synchronization) runs once per step,
independent of batch size. \\
\textit{Implication:} parallelism belongs at the step level, sharing one forward pass per step amortizes
the fixed dispatch cost across a batch.
\end{tcolorbox}

Building on finding (3), we show that independent per-request batching pays the dispatch overhead once per
request per step, while step-level sharing pays it once per step for the whole batch (\S\ref{sec:design}).
We also derive a closed form batch timeout rule from measured service time statistics
(\S\ref{sec:scheduling}).
\section{Background}
\label{sec:background}
\subsection{What Serving Characterization Requires}
A serving characterization has to answer three questions: how much request cost varies, what drives that
variation, and whether the variation has a structure a scheduler can use. In this paper, request cost means
end-to-end wall-clock latency per request, measured from tokenization through detokenization
(\S\ref{sec:setup}). This latency has two components, CPU-side dispatch and GPU compute, and
\S\ref{sec:profiling} shows that these two components are not interchangeable for masked diffusion serving. We report cost at the level of a full
request, not per denoising step, except where we look at per-step dispatch overhead directly
(\S\ref{sec:profiling}).

For autoregressive (AR) serving, these three questions are easy to answer. Decoding moves one token at a
time, and the work per token stays about the same across requests. For masked diffusion serving, none of
this holds by default. The denoising path can differ from one request to the next, and how much work a
request needs depends on when and how its tokens become confident enough to unmask.

\begin{figure}[htbp]
\centerline{\includegraphics[width=0.98\linewidth]{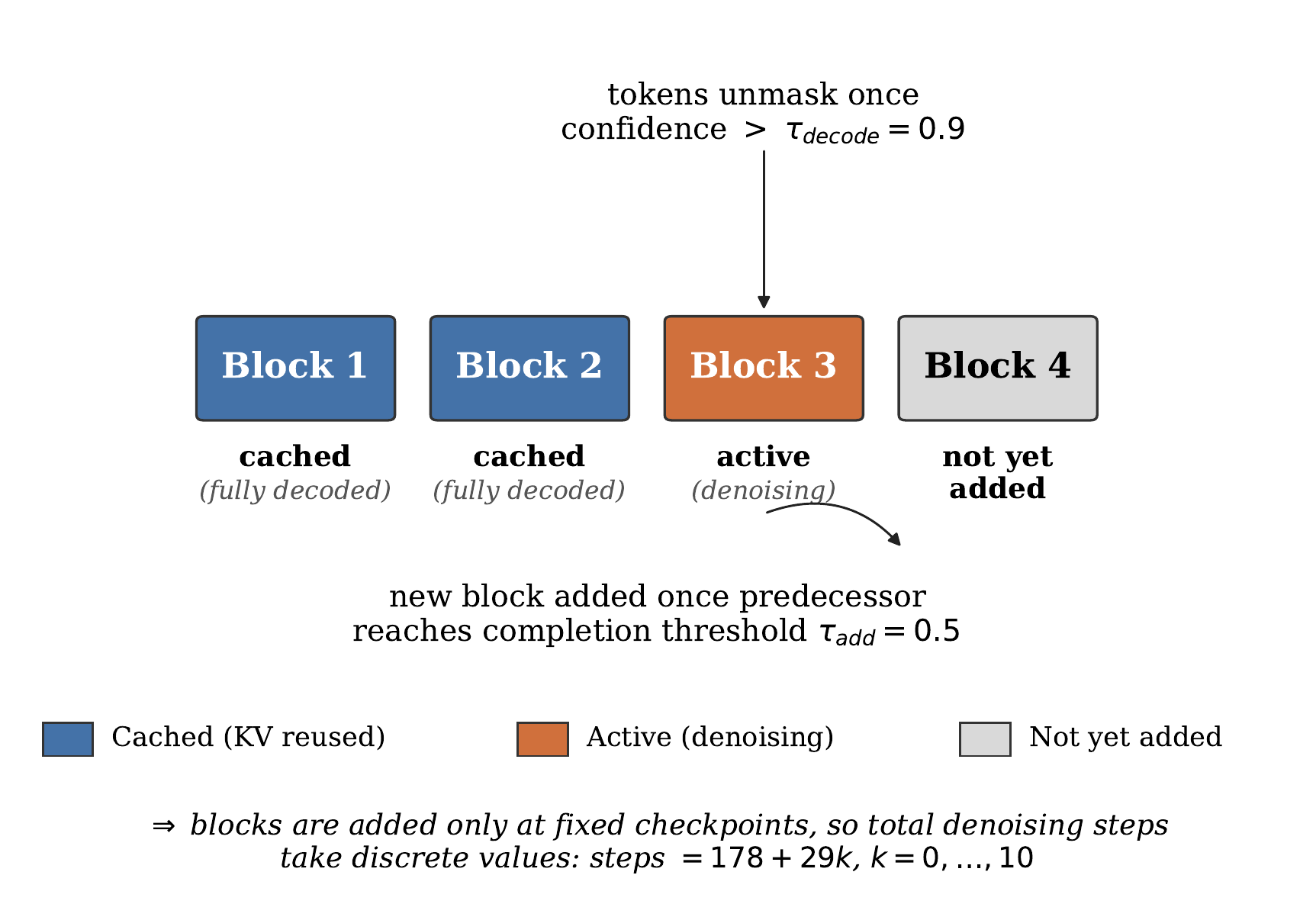}}
\caption{The block-threshold mechanism that induces discrete serving behavior in our test system. A new block
starts only after its predecessor crosses $\tau_{add}$, and tokens in the active block unmask only after
confidence crosses $\tau_{decode}$. We use this mechanism to explain serving-level phenomena rather than
study the mechanism in isolation. }
\label{fig:d2f-mechanism}
\end{figure}

To study this properly, we need a system where we can actually see what is happening inside decoding, and
connect that to the latency we measure. We use D2F \cite{wang2025diffusion} for this. D2F breaks generation
into fixed-size blocks (32 tokens in our setup) and moves blocks forward using two thresholds: a completion
threshold $\tau_{add} = 0.5$, which controls when a new block can start, and a decode threshold
$\tau_{decode} = 0.9$, which controls when a single token gets unmasked. Figure~\ref{fig:d2f-mechanism}
shows this.

We chose D2F because it makes the source of variability visible instead of hidden. A new block can only
start once a fixed checkpoint is crossed, so the number of denoising steps a request needs cannot vary
smoothly. It has to land on one of a small set of fixed values. This is what lets us trace the discrete
difficulty tiers we measure in \S\ref{sec:tiers} back to a real, checkable cause in the decoding process,
rather than leaving them as an unexplained pattern. We follow this same approach throughout the paper: every
serving-level result we report comes with a concrete mechanism behind it.

\subsection{Experimental Setup}
\label{sec:setup}
All experiments use LLaDA-8B-Instruct \cite{nie2026large} with the D2F LoRA adapter, run
on a single NVIDIA H200 GPU in bfloat16. We evaluate on GSM8K~\cite{cobbe2021training} using the model's chat template, and on HumanEval
for the cross-task validation described in \S III-E. We use PyTorch 2.12.1
(CUDA 13.0), transformers 4.49.0, and peft 0.19.1. All generation uses
sampling temperature 0.2. Each experiment sets its own explicit random
seed rather than one shared global seed; the specific values are listed
per experiment in Appendix A.

We measure end-to-end latency from tokenization through detokenization, and we place
\texttt{torch.cuda.synchronize()} around every timed region so that asynchronous GPU dispatch does not
distort the measurement. The scheduling latency percentiles reported later in the paper come from a Monte
Carlo simulation run over real measured service times; every other number we report is a direct GPU
measurement.
\section{Serving Characterization}
\label{sec:characterization}

\subsection{Difficulty is Discrete, Not Continuous}
\label{sec:tiers}
We start by asking a basic question: how much does the cost of a request actually vary, and does that
variation have any shape a scheduler could use? To answer this, we run 100 GSM8K requests through the
system and look at denoising-step behavior. An initial run of 64 requests already showed a clear 11-level
pattern, and we confirmed it again at 100 requests. All numbers below use the full 100-request sample.

\begin{figure}[htbp]
\centerline{\includegraphics[width=0.95\linewidth]{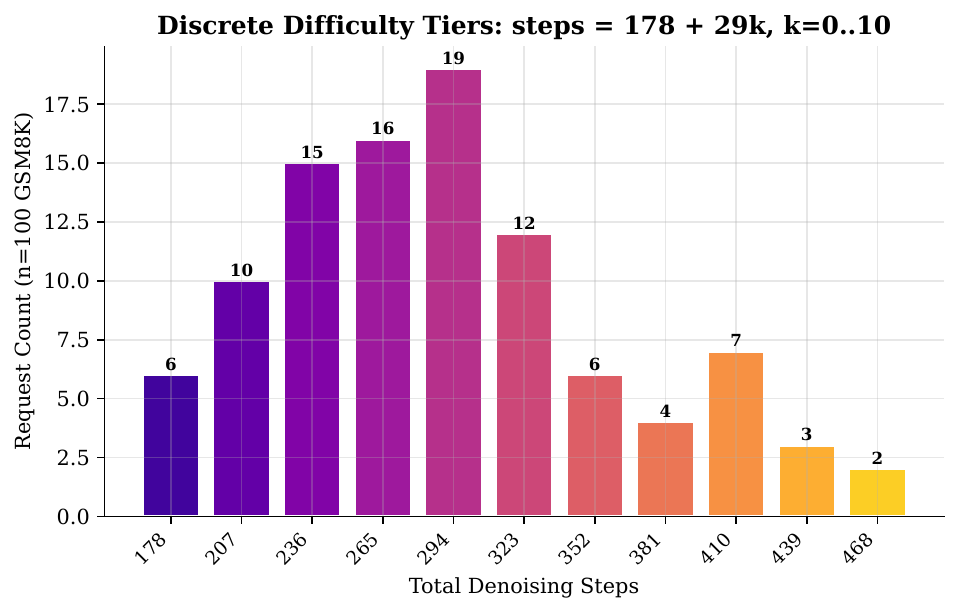}}
\caption{Denoising step counts for 100 GSM8K requests fall into exactly 11 discrete levels, spaced 29 steps
apart.}
\label{fig:tiers}
\end{figure}

Under our fixed decoding thresholds ($\tau_{add}=0.5$, $\tau_{decode}=0.9$), the number of denoising steps a
request needs takes exactly 11 values: 178, 207, 236, and so on up to 468, each one 29 steps apart
(Fig.~\ref{fig:tiers}). This is not noise. It comes directly from the block-addition rule in
\S\ref{sec:background}: a new block can only start once the previous block crosses a fixed completion point.

\begin{figure}[htbp]
\centerline{\includegraphics[width=0.95\linewidth]{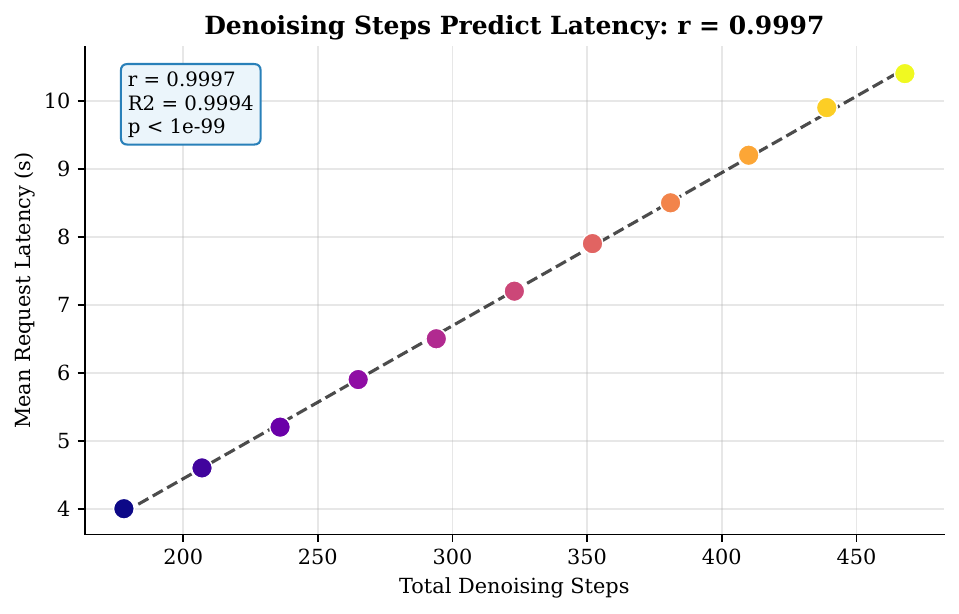}}
\caption{Denoising step count predicts end-to-end latency across the 11 discrete difficulty levels
($r=0.9997$).}
\label{fig:steplatency}
\end{figure}

This structure turns out to matter a great deal for serving, because step count predicts end-to-end latency
almost perfectly (Fig.~\ref{fig:steplatency}): $r = 0.9997$ ($R^2 = 0.9994$, $p < 10^{-99}$). If a system
could learn the step count at admission time, scheduling would become a simple choice among 11 known
outcomes, rather than the much harder problem of predicting a continuous service time. Whether that step
count can actually be learned in advance is the question we take up next.

\begin{tcolorbox}[
  colback=gray!5, colframe=gray!60, boxrule=0.5pt, arc=1pt,
  left=6pt, right=6pt, top=4pt, bottom=4pt
]
\textbf{Observation 1.} A discrete cost structure like this turns scheduling into a classification problem
over 11 known outcomes, not an estimation problem (\S\ref{sec:design}, \S\ref{sec:scheduling}).
\end{tcolorbox}

\subsection{High-Variance Serving Workload}
\begin{figure}[htbp]
\centerline{\includegraphics[width=0.95\linewidth]{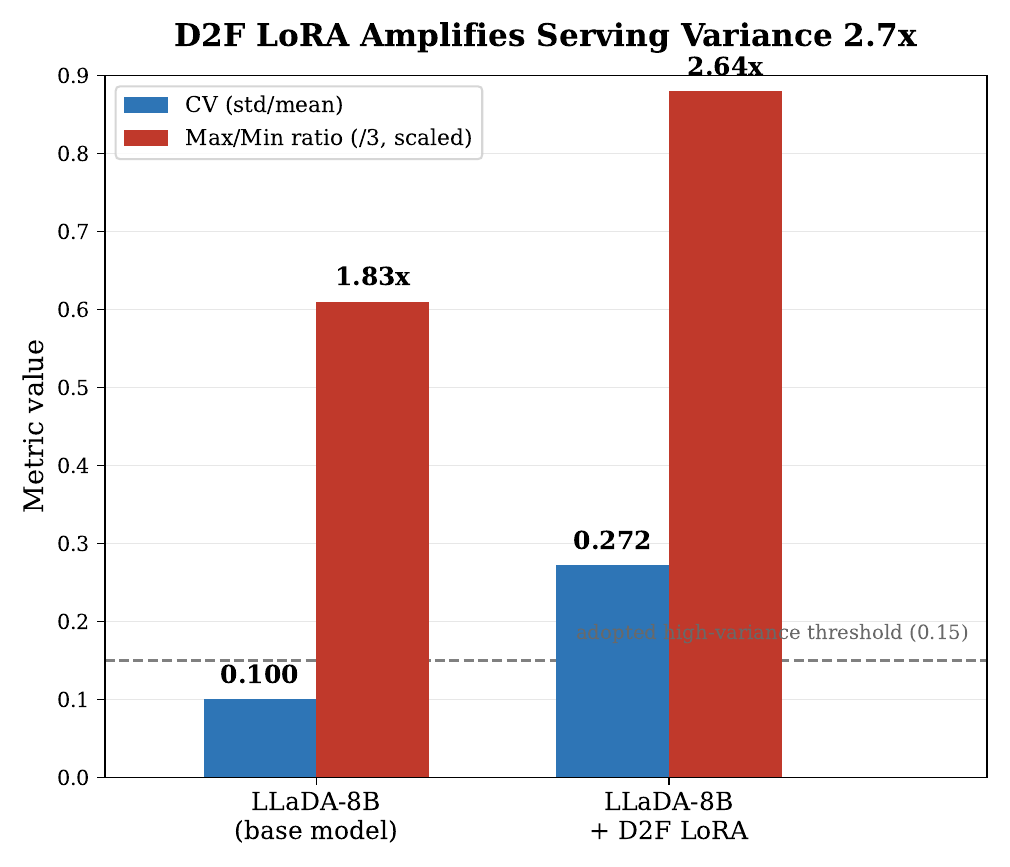}}
\caption{D2F increases serving-latency variance by roughly 2.7$\times$ relative to the base model
($CV = 0.272$ vs. $CV = 0.100$) on identical GSM8K prompts under a 512-token generation ceiling.}
\label{fig:variance}
\end{figure}

Discrete step tiers only matter for scheduling if the resulting latency spread is large enough to care
about, so we check that next. Under D2F, GSM8K prompts show real latency variability: $CV = 0.272$ under a
512-token generation ceiling(Fig.~\ref{fig:variance}), with a max-to-min latency ratio of $2.64\times$. This sits well above
$CV = 0.15$, which we use as a practical threshold for high variance in this paper; 
CV estimates shift somewhat across separate
measurement passes in this paper, ranging from 0.256 to 0.283 across \S\ref{sec:tiers}, Table~\ref{tab:genlen},
and Appendix~B. This is consistent with normal sampling variability at $n=100$, so we report each value as
measured in its own experiment rather than forcing them to a single number.

\begin{tcolorbox}[
  colback=gray!5, colframe=gray!60, boxrule=0.5pt, arc=1pt,
  left=6pt, right=6pt, top=4pt, bottom=4pt
]
\textbf{Observation 2:} The variance comes from D2F's decoding path, not from LoRA adapters in general, since
the base model on the same prompts shows much lower variance.
\end{tcolorbox}

\subsection{Difficulty Cannot Be Predicted Before Generation Starts}
\begin{figure}[htbp]
\centerline{\includegraphics[width=0.95\linewidth]{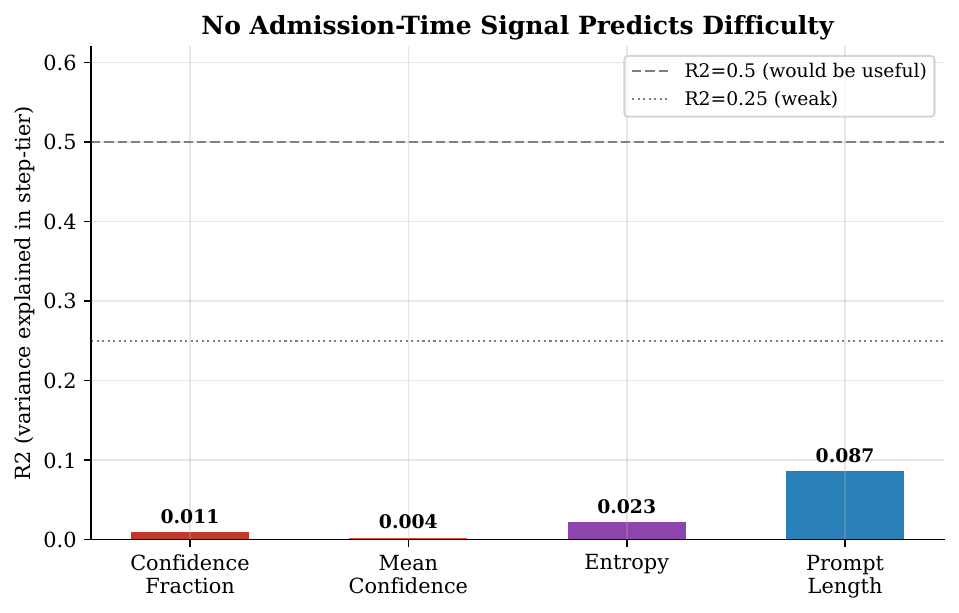}}
\caption{None of the tested early signals predict the eventual difficulty level; all $R^2$ values remain far
below the $R^2=0.5$ heuristic we use as a scheduling-usefulness floor.}
\label{fig:probe}
\end{figure}

Given that difficulty is discrete and highly variable, the natural next step is to try predicting it early
and use that for routing or batching. We tested four cheap signals available right after the first
denoising step: the fraction of tokens past the confidence threshold, average token confidence, output
entropy, and prompt length. Figure~\ref{fig:probe} shows the result: none of the four signals work. Their $R^2$ values against the eventual
difficulty level are 0.011, 0.004, 0.023, and 0.150, all far
below the $R^2 = 0.5$ floor we treat as a practical, self-chosen
minimum: below this point, a signal explains less than half the
variance in the eventual tier, which is too little to act on for
admission-time routing.

The failure is structural, not a weakness in these specific signals. After one denoising step, every request
still has all 32 positions of its first block fully masked, regardless of how hard the problem is. An
early-step signal mostly reflects the model's general prior, not anything about that specific problem.
Difficulty depends on the denoising path a request takes, and that path is not visible before it unfolds.
So the discreteness we found earlier does not make scheduling easier by itself: 11 known outcomes are no
help if none of them can be guessed in advance.

\begin{tcolorbox}[
  colback=gray!5, colframe=gray!60, boxrule=0.5pt, arc=1pt,
  left=6pt, right=6pt, top=4pt, bottom=4pt
]
\textbf{Observation 3:} The failure is structural, not signal-specific, so better features are unlikely to
help. Scheduling has to plan around unpredictability, not by trying to remove it. 
\end{tcolorbox}

\subsection{Short Generation Budgets Hide the Real Variance}
\label{sec:truncation}
\begin{figure}[htbp]
\centerline{\includegraphics[width=0.98\linewidth]{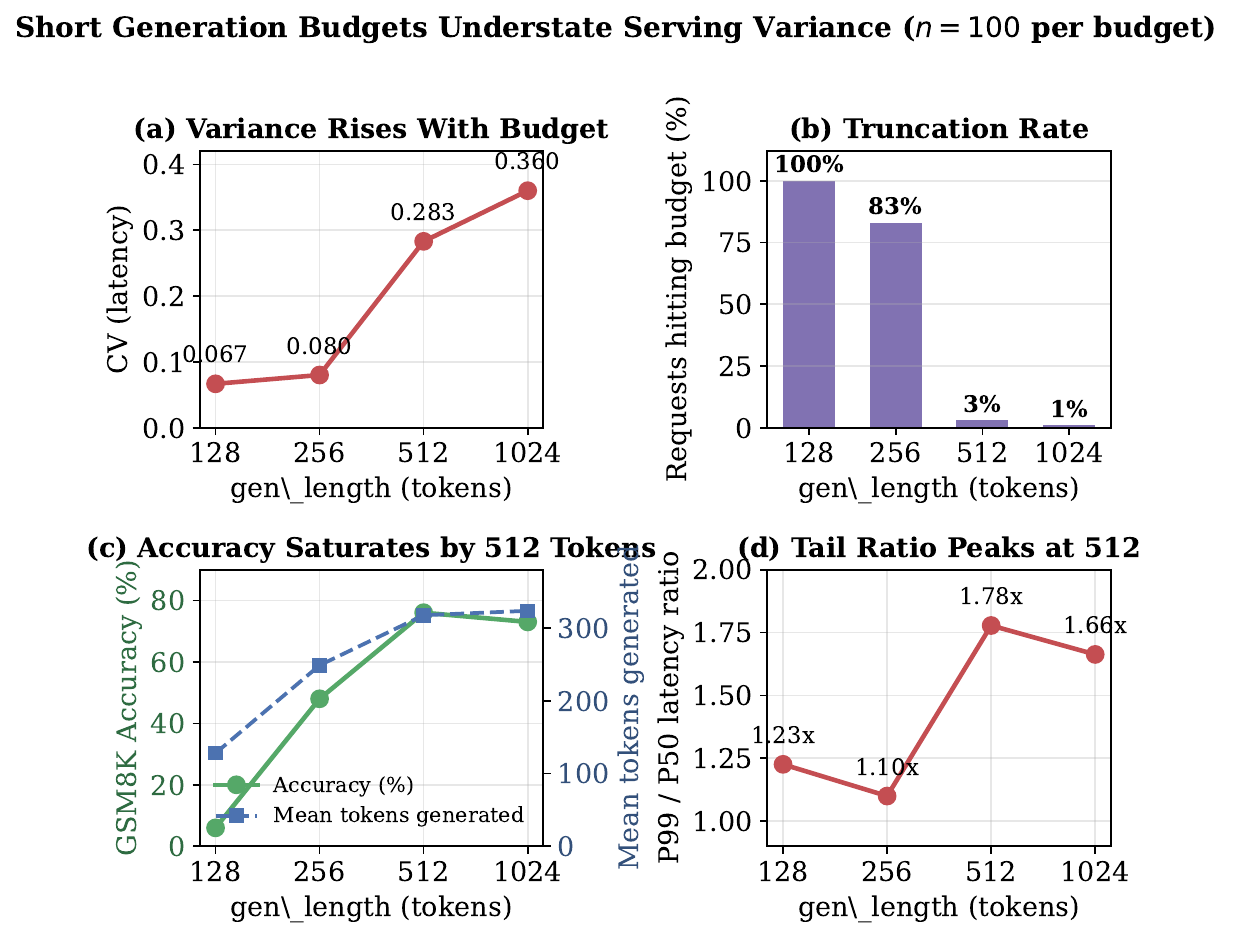}}
\caption{Short generation budgets systematically understate serving variance (100 GSM8K requests per budget).
(a) CV rises with budget, with the sharpest increase between 256 and 512 tokens once truncation no longer
masks variance. (b) Truncation falls from 100\% to 1\%. (c) Accuracy saturates by 512 tokens.
(d) Tail-to-median latency peaks at 512 tokens and decreases slightly by 1024 once typical completion length
is consistently reached.}
\label{fig:truncation}
\end{figure}

The variance numbers above were all measured at a single generation budget, so a natural question is whether
that budget choice was hiding or exaggerating the real picture. To check, we repeat the characterization at
four budgets (128, 256, 512, and 1024 tokens), running 100 GSM8K requests at each budget with single-request
inference. Latency variance rises steadily as the budget grows, with the biggest jump happening once
truncation stops dominating the results (Fig.~\ref{fig:truncation}, Table~\ref{tab:genlen}).

At a 128-token budget, every request gets truncated (100\%), and accuracy is only 6.0\%. At 256 tokens, 83\%
of requests are still truncated and accuracy reaches 48.0\%. By 512 tokens, truncation almost disappears
(3\%) and accuracy reaches 76.0\%; results look similar at 1024 tokens (1\% truncation, 73.0\% accuracy). On
this workload, the model needs roughly 320 tokens on average to finish a GSM8K problem (317.4 tokens at the
512-token budget and 323.5 at the 1024-token budget; Table~\ref{tab:genlen}), so extra budget past that adds
little to accuracy. Still, CV keeps rising between 512 and 1024 tokens (0.283 to 0.360; Table~\ref{tab:genlen}),
which tells us that variance has not fully settled down within the budget range we tested, even past the
point where generation naturally completes.

\begin{table}[htbp]
\caption{Generation-length scaling (100 GSM8K requests per budget)}
\label{tab:genlen}
\begin{center}
\begin{tabular}{@{}lcccccc@{}}
\toprule
\textbf{gen\_len} & \textbf{CV} & \textbf{P50} & \textbf{P90} & \textbf{P99} & \textbf{Trunc.} & \textbf{Acc.} \\
\midrule
128  & 0.067 & 2.84s & 3.12s & 3.48s  & 100\% & 6.0\%  \\
256  & 0.080 & 5.38s & 5.70s & 5.92s  & 83\%  & 48.0\% \\
512  & 0.283 & 6.46s & 9.72s & 11.48s & 3\%   & 76.0\% \\
1024 & 0.360 & 6.41s & 9.05s & 10.66s & 1\%   & 73.0\% \\
\bottomrule
\end{tabular}
\end{center}
\end{table}

Any dLLM serving benchmark run below a task's natural completion length is mostly measuring truncation
behavior, not real serving behavior. The high-variance result reported above ($CV = 0.272$) was measured
separately, under a 512-token ceiling close to the natural completion point identified here. By contrast,
the 128- and 256-token results in Table~\ref{tab:genlen} are directly shaped by truncation.

\begin{tcolorbox}[
  colback=gray!5, colframe=gray!60, boxrule=0.5pt, arc=1pt,
  left=6pt, right=6pt, top=4pt, bottom=4pt
]
\textbf{Observation 4:} This is a benchmarking finding as much as a model finding. A short generation budget
can make a workload look far more predictable than it really is.
\end{tcolorbox}

\subsection{Cross-Task Generalization: HumanEval}
\label{sec:humaneval}
Everything so far was measured on GSM8K math problems alone, which raises an obvious concern: are these
patterns specific to math reasoning, or do they hold more broadly? To check, we repeated the block-size
sweep and the variance characterization on 64 HumanEval code-completion problems. Code generation is also
directly relevant to HPC settings, where LLM-assisted development is an increasingly common facility
workload. Two results held up cleanly.

First, block size behaves purely as a serving-granularity setting; it does not, on its own, determine
difficulty or latency. Within a single request, latency stays almost constant across block sizes
$\{16, 32, 64, 128\}$ (CV $= 0.0096$ on HumanEval), while latency across different requests varies far more
(CV $= 0.345$). That is roughly a 36$\times$ gap, and it matches the roughly 36$\times$ gap we found on
GSM8K almost exactly (within-request CV $0.0073$ versus across-request CV $0.260$).

Second, HumanEval turns out to be, if anything, a harder serving workload than GSM8K. Its CV (0.343 to
0.346) is higher than GSM8K's (0.256 to 0.263), and its max-to-min latency ratio (3.99 to 4.09$\times$) is
higher too (2.60 to 2.70$\times$ on GSM8K). Prompt length correlates a bit more strongly with difficulty
here ($r = 0.648$, $R^2 = 0.42$) than on GSM8K ($r = 0.404$), but it remains far from useful for predicting
difficulty at admission time.

We take the matching roughly 36$\times$ ratio across two very different tasks as consistent, though limited,
evidence that block-size invariance is a general property of D2F generation rather than something specific
to GSM8K. It also suggests our GSM8K variance numbers are, if anything, on the conservative side compared to
other task types. Together, these five findings, discrete tiers, real variance, unpredictability, budget sensitivity, and
cross-task consistency, motivate the serving design we propose in \S\ref{sec:design} and the scheduling rule
we derive in \S\ref{sec:scheduling}.

\begin{tcolorbox}[
  colback=gray!5, colframe=gray!60, boxrule=0.5pt, arc=1pt,
  left=6pt, right=6pt, top=4pt, bottom=4pt
]
\textbf{Observation 5:} The core pattern, latency varies little within a request but a lot across requests,
holds on both benchmarks, with a nearly identical ratio ($\sim$36$\times$) on GSM8K and HumanEval. 
\end{tcolorbox}
\section{System Design: Step-Level Dispatch Cost Scaling}
\label{sec:design}

\subsection{Synchronized Batching}
The serving characterization results point to a design question: if request cost cannot be predicted in
advance, how should a system actually batch requests together? We evaluate two strategies to answer this,
and the difference between them turns out to be large: synchronized batching reaches 16.0$\times$ the
throughput of per-request dispatch at batch size 16 (\S\ref{sec:design}-A, below). The rest of this section
explains what these two strategies are and why the gap between them is this big.

In \emph{synchronized batching}, all requests in a batch are padded to a common shape, and the system runs
one shared forward pass per denoising step for the whole batch. After that shared pass, each request still
samples its own tokens independently. We compare this against a per-request dispatch baseline, where each
request gets its own independent forward pass at every step.

This baseline is not the same thing as continuous batching as implemented in vLLM~\cite{kwon2023efficient}
and Orca~\cite{yu2022orca}, which already share one forward pass across active requests at every iteration.
Their innovation is admission and eviction at iteration boundaries, not per-request dispatch. We isolate the
per-request-dispatch condition here specifically to measure the CPU-overhead effect described in
\S\ref{sec:profiling}. This comparison isolates dispatch overhead. The slot-based design we
describe below (\S\ref{sec:scaling}) would be a more representative comparison; we
propose building it as future work.

\begin{tcolorbox}[
  colback=gray!5, colframe=gray!60, boxrule=0.5pt, arc=1pt,
  left=6pt, right=6pt, top=4pt, bottom=4pt
]
\textbf{Observation 6:} Per-request dispatch is a useful baseline for isolating dispatch cost, but it is not
the strongest baseline a real system would use.
\end{tcolorbox}

\begin{figure}[htbp]
\centerline{\includegraphics[width=0.95\linewidth]{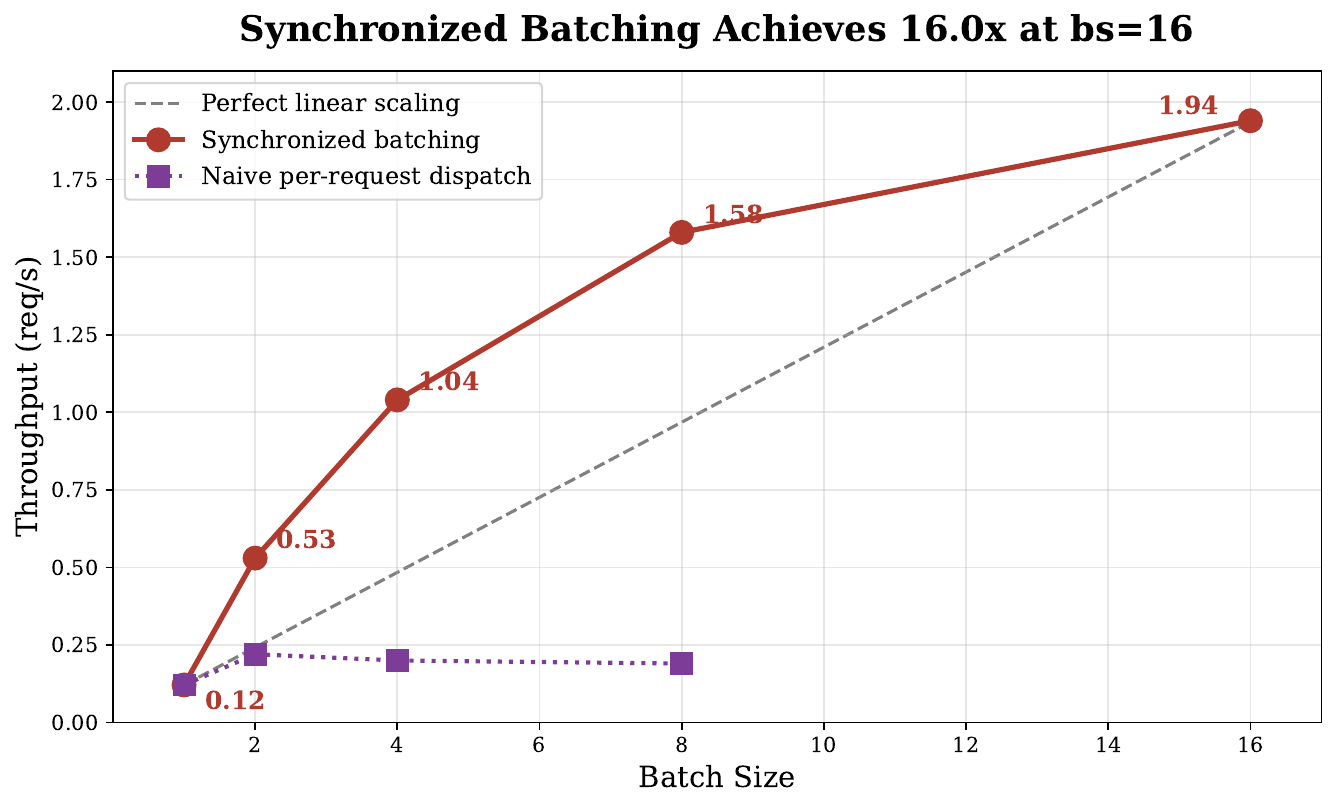}}
\caption{Synchronized batching scales close to linearly up to batch size 16 (16.0$\times$ over single-request
throughput). In contrast, naive per-request dispatch (one forward pass per request) levels off far below
linear scaling.}
\label{fig:sync}
\end{figure}

Synchronized batching reaches 1.939 requests/s at batch size 16, a 16.0$\times$ improvement over
single-request throughput (0.121 requests/s). Throughput scales close to linearly with batch size, because
requests share the dominant attention computation at every denoising step (Fig.~\ref{fig:sync}).

\subsection{Step-Level Dispatch Cost Scaling}
\label{sec:scaling}
\begin{figure}[htbp]
\centerline{\includegraphics[width=0.95\linewidth]{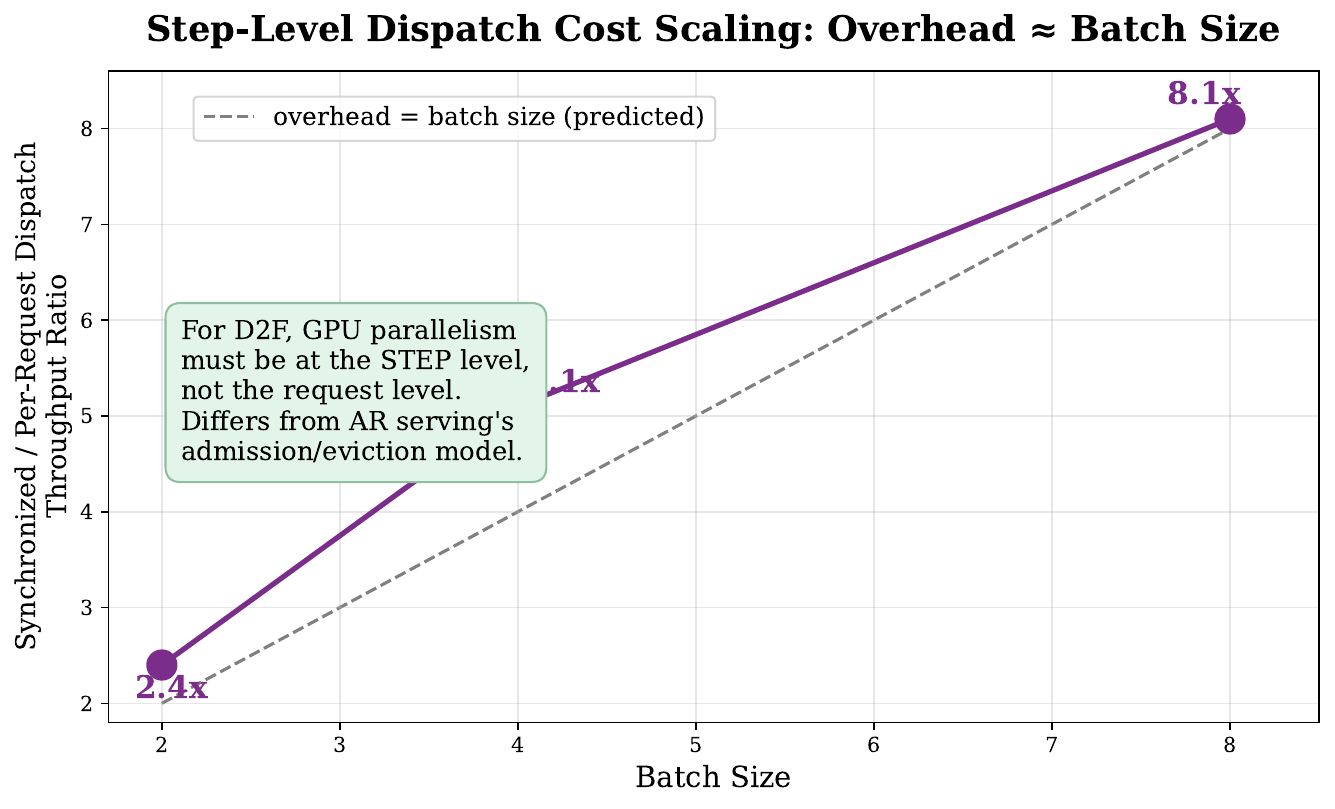}}
\caption{The overhead of naive per-request dispatch relative to synchronized batching tracks batch size, consistent
with one forward pass per request versus one forward pass per step for the batch.}
\label{fig:law}
\end{figure}

The gap between synchronized batching and per-request dispatch is not an implementation quirk; it comes from
how masked diffusion decoding actually computes each step. Here, $B$ denotes the batch size: the number of
requests advanced together in a single synchronized batch, or handled concurrently under per-request
dispatch. We measure the extra overhead that per-request dispatch carries relative to synchronized batching.
This overhead tracks $B$ closely: 2.4$\times$ at $B=2$, 5.1$\times$ at $B=4$, and 8.1$\times$ at $B=8$
(Fig.~\ref{fig:law}). At smaller batch sizes, the overhead sits within 20 to 27\% of $B$ itself, and by
$B=8$ it matches $B$ almost exactly.

This overhead scaling follows directly from how masked diffusion decoding computes each step: advancing $B$
requests independently takes $B$ forward passes per denoising step, while advancing them jointly takes one.
We measured this pattern directly for D2F and traced it to its mechanistic cause through kernel-level
profiling (\S\ref{sec:profiling}), showing that the per-block control loop, not GPU compute, is what
batching actually amortizes. This measurement and its mechanistic explanation are our contribution here,
distinct from the batch-size scaling itself, which follows from the per-request-dispatch baseline by
construction and would not appear under a shared-pass design such as a slot-based system
(\S\ref{design_principle}) or a vLLM-style batcher adapted to per-step admission.

Systems such as dLLM-Serve~\cite{fan2025memoryfootprint} and Sangam~\cite{kedia2026sangam} also explore
step-level batching for the LLaDA family. What we add is the direct, D2F-specific measurement of this
overhead and its connection to a concrete mechanistic cause. This differs from AR batching, where sharing
the forward pass is already standard practice; the difference here is that masked diffusion's per-block
dispatch cost cannot be amortized through admission and eviction alone, the way it can under AR's simpler
next-token dependency.

\textbf{Design principle.}\label{design_principle} In masked diffusion serving, step-level sharing is built
into the cost itself. One forward pass can advance every request in a batch through a denoising step. But
advancing requests independently multiplies that cost by the batch size.

Because of this, a serving system needs a way to let requests join and leave over time. At the same time, it
needs to keep synchronized batching's throughput. Doing both means combining shared forward passes with
per-request completion.

One way to do this is a slot-based design. Keep a fixed number of GPU slots. Move all active slots forward
together through one shared forward pass at each step. Refill any slot from a queue as soon as its request
finishes.

This slot-based design is, in practice, the fair comparison point for synchronized batching. It shares the
forward pass, like synchronized batching does, but it also lets requests join and leave, which the
per-request-dispatch baseline we used above does not. We propose this as  future work.


\begin{tcolorbox}[
  colback=gray!5, colframe=gray!60, boxrule=0.5pt, arc=1pt,
  left=6pt, right=6pt, top=4pt, bottom=4pt
]
\textbf{Observation 7:} The $B\times$ overhead is a consequence of the baseline we chose, not a discovery
about masked diffusion itself. The real contribution is measuring where that cost comes from.
\end{tcolorbox}

\subsection{Where the Time Actually Goes}
\label{sec:profiling}
To explain the throughput results from the previous two subsections, the
16.0× improvement from synchronized batching and the overhead that scales
with batch size B, we profile a single request on the H200 at our
standard 512-token generation budget (\S II-B). We use \texttt{torch.profiler}, and we check the breakdown by summing raw kernel-execution durations
from the exported Chrome trace. We treat this trace-summation value as ground truth. It matches PyTorch's
own printed \emph{Self CUDA time total} to three decimal places. We rely on this trace-summation method specifically because it matches PyTorch's own reporting; the
built-in \texttt{key\_averages()} summary does not, returning exactly twice the true value in our profiler
build (Appendix~A).

\begin{figure}[htbp]
\centerline{\includegraphics[width=0.8\linewidth]{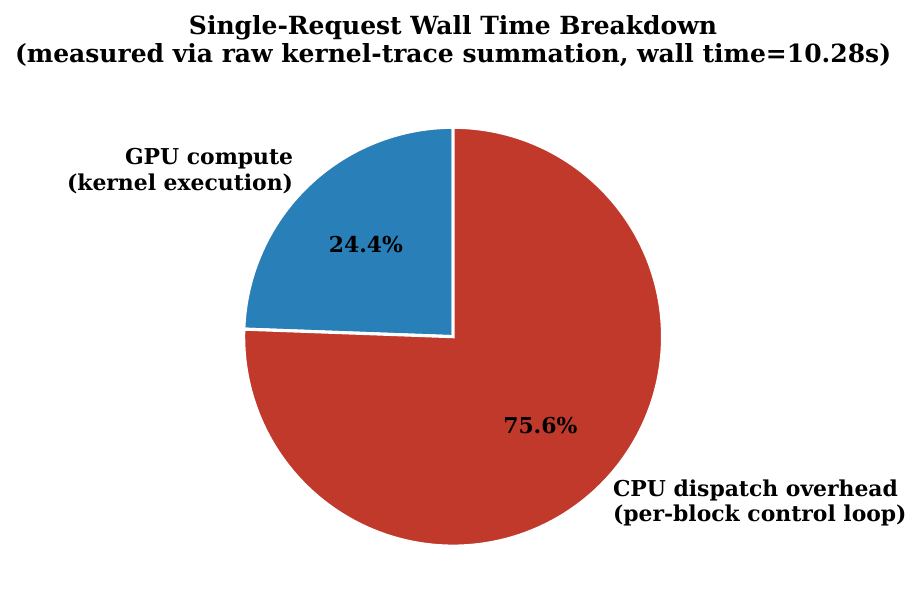}}
\caption{Single-request wall-clock breakdown from kernel-trace summation: 76\% CPU-side dispatch overhead in
the per-block control loop versus 24\% GPU kernel time.}
\label{fig:profile}
\end{figure}

A full generation takes 10.28~s wall-clock. Only 2.511~s (24.4\%) of that is GPU kernel time. The remaining
75.6\% is CPU-side overhead in the per-block control loop: tracking block state, building attention masks,
checking thresholds, and issuing synchronization between forward passes (Fig.~\ref{fig:profile}).

This breakdown explains why batching helps. The per-block control loop runs once per denoising step, no
matter how many requests are being advanced. Synchronized batching pays this cost once per step for the
whole batch. Per-request dispatch pays it once per step for every single request. So batching mainly helps
by amortizing fixed CPU dispatch overhead, not by improving GPU utilization the way typical AR serving does.

This host-side bottleneck also matters for HPC-AI infrastructure. dLLM serving is currently limited by CPU
dispatch, not by GPU compute. Because of this, it should benefit a lot from tightly coupled CPU-GPU designs,
such as GH200-class superchips with fast, low-latency links between host and device. It should also benefit
from dispatch-elimination techniques such as CUDA Graphs, which capture the control loop as a replayable
graph and remove per-step launch overhead. A result that looks like a single-GPU profiling detail is really
an infrastructure statement: the per-block control loop, not the accelerator, is the current ceiling on dLLM
serving throughput. Closing that gap is as much an architecture and systems-integration problem as a
modeling one.

\begin{tcolorbox}[
  colback=gray!5, colframe=gray!60, boxrule=0.5pt, arc=1pt,
  left=6pt, right=6pt, top=4pt, bottom=4pt
]
\textbf{Observation 8:} The bottleneck is CPU dispatch, not the GPU. This points serving optimization toward
architecture and systems integration, not just kernel-level speedups.
\end{tcolorbox}

\subsection{Forward-Pass Launch Overhead: A CUDA Graphs Micro-Experiment}
To break the dispatch overhead from \S\ref{sec:design}-B down further, we isolate one forward-pass call and
capture it with \texttt{torch.cuda.CUDAGraph}. This is the first call of a generation, at the prompt-only
shape, with no KV cache yet. Capture cuts this call's CPU dispatch time by 48.6\% (19.4~ms to 10.0~ms). We
also confirmed the output stays numerically identical: the logits match exactly, with a maximum absolute
difference of $0.0$.

Next, we ask how much this saves across a full generation. We instrument
\texttt{\_generate\_block\_single} to count forward-pass calls for the same request profiled in
\S\ref{sec:profiling}: 235 calls. We confirmed this count independently through
\texttt{aten::embedding} operation counts in the raw kernel trace (235 out of 235, an exact match). The
64-to-1 ratio of attention-op to embedding-op counts also confirms the model has 64 transformer layers.

For similar savings across all 235 calls, across the different KV-cache shapes later
calls use, graph capture of the forward pass alone would remove
235 × 9.43ms = 2.22s of the 7.77s total dispatch overhead. That is 28.5\%. The remaining
71.5\% comes from per-block Python control flow: mask construction, threshold evaluation, and block-state
bookkeeping outside the forward call itself. This measurement used a 512-token generation budget.

\begin{tcolorbox}[
  colback=gray!5, colframe=gray!60, boxrule=0.5pt, arc=1pt,
  left=6pt, right=6pt, top=4pt, bottom=4pt
]
\textbf{Engineering implication.} CUDA graph capture is a partial, practical fix already trusted by the HPC
community. It covers roughly a quarter of per-step dispatch overhead, with low risk. The larger remaining
share needs a restructured control loop.
\end{tcolorbox}

\subsection{Evaluation: Does Batching Hurt Quality?}
\label{sec:eval}
\begin{figure}[htbp]
\centerline{\includegraphics[width=0.95\linewidth]{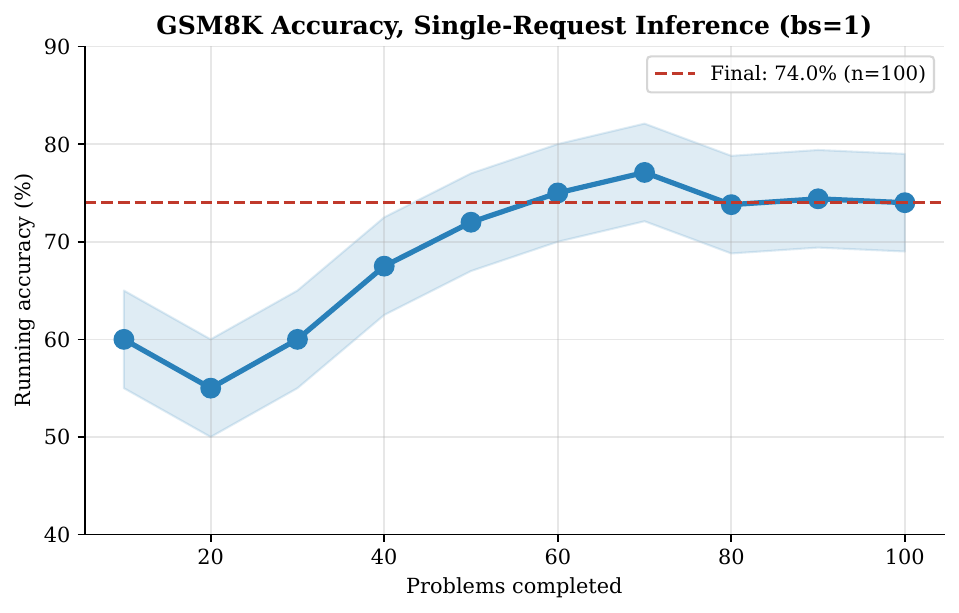}}
\caption{GSM8K exact-match accuracy under single-request inference at \texttt{gen\_length}$=512$ (100 problems).}
\label{fig:accuracy}
\end{figure}

A serving speedup only matters if it does not cost accuracy. Under single-request inference on GSM8K at
\texttt{gen\_length}$=512$, we measure 74.0\% exact-match accuracy (74/100) in this experiment. We measured
76.0\% (76/100) in a separate experiment with $n=100$ (\S\ref{sec:truncation}, Table~\ref{tab:genlen}). Both
use the same model, thresholds, and generation length. The two-point gap comes from normal run-to-run
variation under temperature $>0$ sampling, not a difference in method (Fig.~\ref{fig:accuracy} shows the
74.0\% run).

A closer look at the 24 incorrect responses shows that only 2 (8.3\%) reached the 512-token budget,
consistent with truncation. The remaining 22 completed generation normally but gave a wrong final answer.
So truncation is a minor contributor to the error rate at this generation length, not the main one.

For synchronized batching, we give a structural argument for why output quality should not depend on batch
size. We state the assumptions plainly. The argument needs three things to hold. Each request's attention
mask and positional encoding stay the same under padding, no matter the batch size. Each request samples
tokens from its own independent RNG stream. And the batched attention and sampling kernels are numerically
equivalent, not just statistically similar, to their single-request versions.


\begin{tcolorbox}[
  colback=gray!5, colframe=gray!60, boxrule=0.5pt, arc=1pt,
  left=6pt, right=6pt, top=4pt, bottom=4pt
]
\textbf{Observation 9:} Single-request accuracy holds steady across two runs (74--76\%).
\end{tcolorbox}
\section{Scheduling: A Batch-Timeout Stability Rule}
\label{sec:scheduling}
A serving system has to decide how long to wait for a batch
to fill before it sends it off. Waiting too long adds idle time
and delay for each request. Sending too early wastes forward
passes on batches that are not full. We look at this trade-off for
fixed-fill synchronized batching under Poisson arrivals. This is
a direct application of bulk-service queueing results~\cite{deb1973optimal} to the
step-level dispatch cost of masked diffusion serving. Table~\ref{tab:notation} summarizes the notation we
use throughout this section.

\begin{table}[htbp]
\caption{Notation used in the scheduling model (\S\ref{sec:scheduling})}
\label{tab:notation}
\begin{center}
\begin{tabular}{@{}cl@{}}
\toprule
\textbf{Symbol} & \textbf{Meaning} \\
\midrule
$B$      & Fixed batch size (synchronized batching), $B=8$ throughout \\
$\lambda$ & Request arrival rate (requests/s), Poisson process \\
$S$      & Mean service time per batch, measured as $S=8.74$~s \\
$T$      & Batch timeout: max wait before dispatching an underfull batch \\
$\rho$   & Utilization: arrival rate / service capacity ($\rho \to 0$ idle, $\rho \to 1$ saturated) \\
$T_{\min}$ & Minimum safe timeout; if $T < T_{\min}$, the queue becomes unstable \\
\bottomrule
\end{tabular}
\end{center}
\end{table}

Our rule is grounded in measured service times, not an idealized distribution. We use $n=8$ measured batch
service times at batch size $B=8$ (mean $S=8.74$~s, $CV=0.266$). We check this against a 20{,}000-request
Monte Carlo simulation that resamples from the measured service-time distribution. Table~\ref{tab:sched}
shows the resulting latency percentiles.

\begin{figure}[htbp]
\centerline{\includegraphics[width=0.95\linewidth]{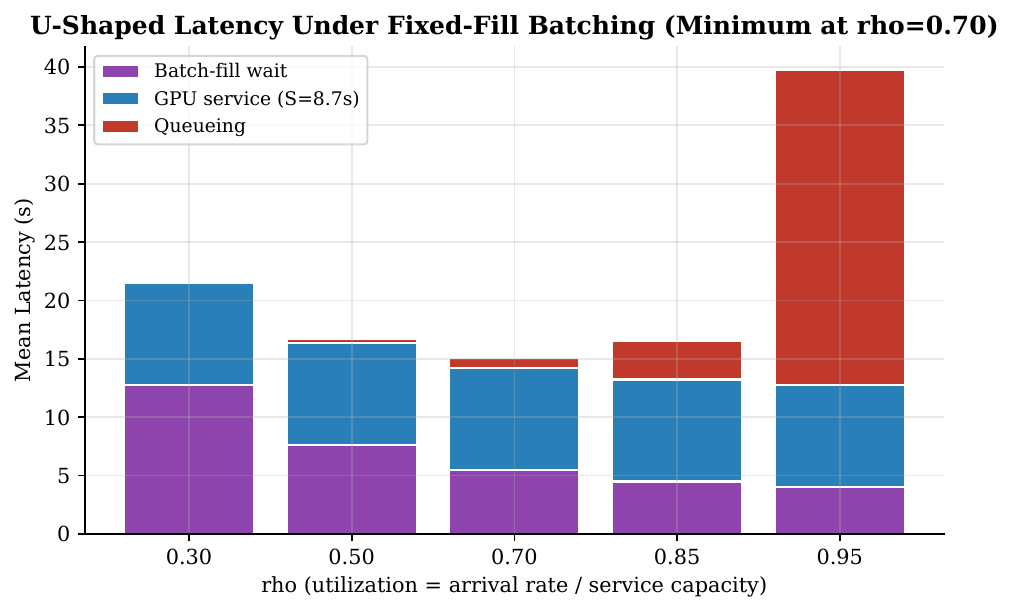}}
\caption{Mean latency under fixed-fill synchronized batching is U-shaped in utilization $\rho$: batch-fill
wait dominates at low $\rho$, while queueing behind other batches dominates at high $\rho$.}
\label{fig:ushape}
\end{figure}

Under a fixed-fill policy, mean latency follows a U-shape as load increases (Fig.~\ref{fig:ushape}). We
define $\rho$ as the ratio of the arrival rate to the system's service capacity. For fixed-fill batches at
full size, capacity is $B/S$ requests per second, about $0.915$ req/s here; a partial batch still costs one
full forward pass, so it lowers effective capacity below this figure. So $\rho \to 0$ means the system is
mostly idle, and $\rho \to 1$ means it is close to saturated.

At low $\rho$, latency is mostly batch-fill delay: requests wait for more arrivals before the batch goes
out. For batch size $B$, the mean fill-wait is $(B-1)/(2\lambda)$, which is $3.5/\lambda$ when $B=8$. At high
$\rho$, latency comes mostly from queueing behind earlier batches, not from waiting for the current one to
fill. Across both effects, the lowest mean latency in our simulation happens near $\rho \approx 0.70$.

This U-shape leads to a simple stability rule. Recall from Table~\ref{tab:notation} that $T$ is the batch
timeout, the longest the system waits before sending a batch even if it is not full, and $S$ is the mean
service time per batch, measured here as $8.74$s. Because a partial batch still costs one full forward pass
under step-level parallelism, the system stays stable only if the expected number of requests collected per
batch keeps up with the rate at which requests arrive, $\lambda$. Formally, this requires
$E[\text{batch size}](T,\lambda) = \lambda S$: the average batch size a timeout $T$ produces, under arrival
rate $\lambda$, must match the arrival rate scaled by how long each batch takes to serve.

When arrivals are sparse within a single timeout window, meaning it is unlikely that many requests arrive
in the same short window, we can approximate the expected batch size as
$E[\text{batch size}](T,\lambda) \approx 1 + \lambda T$. Here, the first arrival is what opens the timeout
window in the first place, and each additional arrival accumulates at rate $\lambda$ over the remaining
time $T$. Substituting this approximation into the stability condition and solving
$1 + \lambda T = \lambda S$ for $T$ gives a simple closed-form bound on the minimum safe timeout,
$T_{\min}$:
\[
T_{\min} = S - \frac{1}{\lambda}.
\]
This bound gets less accurate as $\rho \to 1$, since several arrivals often land in the same window (at
$\rho=0.95$, $\lambda T \approx 6.6$). It also ignores the fixed batch cap $B=8$. At high utilization, the
Monte Carlo results in Table~\ref{tab:sched} are the more reliable estimate.

$T_{\min}$ is the minimum safe timeout. If $T < T_{\min}$, the system sends out too many underfilled batches.
Since each underfilled batch still costs a full forward pass, effective capacity drops below $\lambda$, the
queue grows unstable, and latency diverges.

\begin{table}[htbp]
\caption{Latency percentiles under Poisson arrivals from a Monte Carlo simulation over $n=8$ measured batch
service times (batch size $B=8$)}

\label{tab:sched}
\begin{center}
\begin{tabular}{@{}lcccccc@{}}
\toprule
$\rho$ & $\lambda$ (req/s) & P50 & P90 & P99 & $T_{\min}$ & SLA viol.$^{\mathrm{b}}$ \\
\midrule
0.30 & 0.275 & 19.6s & 36.2s & 52.3s & 5.1s & 33\% \\
0.50 & 0.458 & 15.8s & 25.9s & 34.7s & 6.6s & 12\% \\
0.70 & 0.641 & 14.6s & 22.2s & 29.2s & 7.2s & 5\%  \\
0.85 & 0.778 & 16.1s & 23.9s & 30.9s & 7.5s & 8\%  \\
0.95 & 0.870 & 27.2s & 78.4s & 87.9s & 7.6s & 55\% \\
\bottomrule
\multicolumn{7}{l}{\footnotesize $^{\mathrm{b}}$SLA violation defined as end-to-end latency exceeding 25s.}\\
\multicolumn{7}{l}{\footnotesize $^{\mathrm{a}}$Based on $n=8$ measured service times. A subsequent $n=32$}\\
\multicolumn{7}{l}{\footnotesize replication (\S\ref{sec:limitations}) shifted estimated capacity by 24\%; these}\\
\multicolumn{7}{l}{\footnotesize percentiles should be treated as preliminary pending that larger sample.}
\end{tabular}
\end{center}
\end{table}

\section{Limitations and Future Work}
\label{sec:limitations}
Our experiments cover one architecture, LLaDA (base and D2F), on one GPU, the H200. We tested task
generalization (HumanEval, \S\ref{sec:humaneval}) and configuration generalization (base vs.\ D2F,
\S\ref{sec:characterization}-B) separately, not together; testing both jointly, and on other hardware, is
natural next work.

Our scheduling analysis (\S\ref{sec:scheduling}) uses 8 measured batches. A later 32-batch run shifted the
estimated capacity by 24\%, so a larger sample is needed before treating these percentiles as final. Other
single-run numbers in this paper, including Table~\ref{tab:genlen} and Fig.~\ref{fig:sync}, likely carry
similar uncertainty, since none of our results use repeated trials.

We do not compare against existing dLLM inference engines such as dInfer~\cite{ma2025dinfer}. Our
contribution is workload characterization, not a competing system, but a direct comparison would strengthen
\S\ref{sec:scheduling}'s scheduling rule.

Our tier structure (178 + 29k, \S\ref{sec:tiers}) is measured at one threshold setting only
($\tau_{add}=0.5$, $\tau_{decode}=0.9$); these constants are likely specific to this configuration. 

Confirming quality under batching at sizes 8 and 16, and building a working slot-based design, remain as future works.
\section{Related Work}
\textbf{Diffusion language models and acceleration.} Masked diffusion language models
\cite{austin2021structured,lou2024discrete} generate text by unmasking a sequence in parallel, instead of
writing tokens strictly left to right the way autoregressive (AR) decoding does. Most recent work in this
space has focused on cutting \emph{single-request} latency. Caching methods such as dKV-Cache
\cite{ma2025dkvcache}, Fast-dLLM \cite{wu2025fastdllm}, dLLM-Cache \cite{liu2025dllmcache}, and FlashDLM
\cite{hu2025flashdlm} reuse computation across denoising steps, since diffusion models cannot reuse an
AR-style KV cache directly. Sparse-dLLM \cite{song2025sparsedllm} further cuts memory by evicting low-value
cache entries. Spiffy \cite{agrawal2025spiffy} uses speculative decoding to skip steps. D2F
\cite{wang2025diffusion}, which we build on here, restructures generation into blocks to make step-to-step
reuse possible in the first place.

A smaller group of systems reports batched throughput directly. dInfer \cite{ma2025dinfer} supports batched
inference. Concurrent work, including DiLaServe \cite{dilaserve2026} and Sangam \cite{kedia2026sangam},
studies scheduling and resource partitioning for dLLM serving under concurrent load. What none of this work
provides is a measurement-grounded account of \emph{why} dLLM serving behaves the way it does under
concurrent load: how difficulty is structured, whether it can be predicted, and how sensitive results are to
benchmark choices. That gap is what this paper fills, and it is the foundation the design choices in
\S\ref{sec:design} and \S\ref{sec:scheduling} build on.

\textbf{LLM serving systems.} vLLM \cite{kwon2023efficient} and Orca \cite{yu2022orca} are good examples of
AR serving through continuous batching. This design works well because AR decoding is naturally
request-parallel: each request moves forward one token at a time, using only its own history. This is the
assumption our results push back on. We show in \S\ref{sec:design} that this assumption does not carry over
to masked diffusion decoding. For dLLMs, the more efficient strategy is usually to share computation across
requests at each denoising step, rather than run one forward pass per request the way AR-style dispatch
does.

dLLM-Serve \cite{fan2025memoryfootprint} builds a full serving system for diffusion models. It handles
memory budgeting, scheduling, and sparse attention, so it can support many concurrent requests without
running out of GPU memory. That system assumes a certain understanding of dLLM workload behavior already
exists; our work supplies that understanding directly. We measure, on real hardware, how request difficulty,
latency variance, and batching behavior actually show up in masked diffusion serving, so the two lines of
work are complementary rather than overlapping.

\textbf{Queueing analysis for LLM serving.} Earlier work on AR serving latency, such as Splitwise
\cite{patel2024splitwise}, often treats request service costs as independent and additive. That assumption
does not hold here either: under step-level parallel generation, even a partial batch costs a full forward
pass, so service costs are tied to batch composition, not just to individual requests. Our batch-timeout
stability rule is built to handle exactly this. It parallels the batch-size-dependent service times studied
for GPU inference servers by Cui et al. \cite{cui2021queueing}, and the classical bulk-service control
problem studied by Deb and Serfozo \cite{deb1973optimal}. We extend this bulk-service framework to the
timeout-stability condition that step-level-parallel dLLM decoding creates, rather than claim to have
originated bulk-service queueing analysis itself.
\section{Conclusion}
We present a measurement-driven characterization of masked diffusion language model (dLLM) serving on real
hardware. Our measurements support three findings: (i) request difficulty has discrete structure that is not
predictable at admission time, (ii) short generation budgets systematically understate serving variance due
to truncation, and (iii) CPU-side dispatch overhead, not GPU compute, dominates single-request latency (76\%
of wall-clock time versus 24\% GPU kernel time).

The third finding explains why efficient dLLM serving requires \emph{step-level} parallelism rather than the
request-level parallelism typical of autoregressive (AR) serving systems. The per-block dispatch loop runs
once per denoising step regardless of batch size; sharing a forward pass across a synchronized batch
amortizes this fixed cost, yielding a 16.0$\times$ throughput improvement over advancing requests
independently.

Future work includes
validating quality under batching, building a fully correct slot-based step-sharing design, and further
testing the batch-timeout stability rule.

\bibliographystyle{IEEEtran}
\bibliography{main_ref}

\appendices

\section{Profiler Measurement Methodology}
Our GPU profiling result needed careful handling, because of a reliability problem in our PyTorch profiler
build. When we summed per-event self-device-time using \texttt{prof.key\_averages()}, we got a GPU-compute
estimate exactly twice the value shown in PyTorch's own printed summary (`Self CUDA time total') for the
same profiling run. Separately, our \texttt{record\_function} wrapper marker reported an inflated self-time,
almost equal to the full wall-clock duration, whenever it wrapped a long Python loop containing many
\texttt{torch.cuda.synchronize()} calls. This is a known type of accounting error that happens with
long-lived outer profiler markers. Neither problem went away when we called \texttt{key\_averages()} only
once, which was our first guess at a fix, and neither went away when we simply removed the wrapper marker.

Instead, we found ground truth by summing raw kernel-execution durations straight from the exported Chrome
trace (\texttt{prof.export\_chrome\_trace()}), keeping only events tagged \texttt{cat == 'kernel'}. This
method matched PyTorch's own printed total to three decimal places across repeated runs (2.511s), and it is
the number we report in our profiling result. We mention this in detail because anyone profiling masked
diffusion generation loops will likely hit the same issue. Unlike a normal autoregressive decoding loop, a
diffusion loop has many explicit synchronization points inside a single profiled region, and the standard
aggregate profiler API cannot be trusted in that setting.

All experiments here use LLaDA-8B-Instruct with the D2F LoRA adapter on an NVIDIA H200. GSM8K is the primary
benchmark; HumanEval is used only for the cross-task validation. The scheduling latency percentiles in this
paper come from a Monte Carlo simulation run over real measured service times. Every other reported number
is a direct GPU measurement.

\section{Artifact Description (AD)}
This appendix follows the SC Reproducibility Initiative artifact-description convention and is excluded
from the 4--10 page body limit, per the workshop call for papers.

\subsection{Abstract}
This artifact reproduces the workload characterization, the batching throughput and GPU kernel-trace
profiling, the scheduling simulation, and the correctness evaluation reported in the main paper. All
measurements that need GPU access were collected on a single NVIDIA H200. The scheduling simulation is a
pure-CPU Monte Carlo step that runs on the measured service-time data and needs no GPU access to re-run.

\subsection{Artifact Check-List (Meta-Information)}
\begin{itemize}
\item \textbf{Algorithm:} Block-parallel masked-diffusion decoding (D2F); denoising-step characterization;
synchronized batching vs. per-request dispatch; Monte Carlo M/G/1-style queueing simulation.
\item \textbf{Program:} Python 3.10; PyTorch (CUDA/bfloat16 build); HuggingFace \texttt{transformers} and
\texttt{peft} for LoRA adapter loading.
\item \textbf{Model:} LLaDA-8B-Instruct (base) plus the D2F LoRA adapter, bfloat16 precision.
\item \textbf{Data set:} GSM8K test split as the primary dataset, with HumanEval used for cross-task
validation (\S\ref{sec:humaneval}). Both are standard, publicly available benchmarks.
\item \textbf{Run-time environment:} Linux, a CUDA-enabled PyTorch runtime, single-GPU process. No
multi-GPU or distributed execution is used anywhere in this artifact.
\item \textbf{Hardware:} One NVIDIA H200 GPU for all GPU-bound measurements. No GPU is needed to re-run the
scheduling simulation from the provided service-time data.
\item \textbf{Run-time state:} A single-tenant GPU with no co-located jobs. \texttt{torch.cuda.synchronize()}
brackets every timed region, so timestamps are recorded only once asynchronously dispatched CUDA work has
actually finished.
\item \textbf{Execution:} Command-line Python scripts, one per experiment (workload characterization,
batching throughput, GPU profiling, scheduling simulation, quality evaluation). None of this depends on a
notebook environment.
\item \textbf{Output:} Per-request CSV files with latency, step count, and block count, plus aggregate JSON
summaries with CV, percentiles, and throughput for each experiment. GPU profiling also produces a
Chrome-trace JSON file for kernel-level inspection.
\item \textbf{Experiment workflow:} First, we characterize service time and variance. Second, we measure
batching throughput and profile GPU kernels. Third, we run the scheduling Monte Carlo simulation using the
service-time data from the second step. Fourth, we evaluate GSM8K exact-match accuracy.
\item \textbf{Publicly available:} Model weights and benchmark datasets are publicly available from their
providers. Experiment scripts are available from the authors on request, pending the outcome of double-blind
review.
\end{itemize}

\subsection{Installation}
Install a CUDA-enabled PyTorch build that matches your GPU's driver and CUDA version, along with HuggingFace
\texttt{transformers} and \texttt{peft} for loading the LoRA adapter. Download the LLaDA-8B-Instruct base
weights and the D2F LoRA adapter to local storage. No custom CUDA kernel needs to be compiled. All inference
runs through standard PyTorch and HuggingFace code paths.

\subsection{Experiment Workflow}
\textbf{(1) Characterization.} For each benchmark (GSM8K and HumanEval) and each generation-length budget
(128, 256, 512, and 1024 tokens), run single-request inference over a sample of prompts. Record per-request
latency, denoising step count, and block count, with \texttt{torch.cuda.synchronize()} around each timed
region. Sample sizes range from 32 to 100 requests depending on the experiment, as noted next to each table
in the main paper.

\textbf{(2) Batching and profiling.} Run synchronized batching, where every request in a batch shares one
forward pass per denoising step, and per-request dispatch, where each request gets its own independent
forward pass per step, at batch sizes 1, 2, 4, 8, and 16. Record end-to-end batch latency and compute
throughput for each. Separately, profile a single-request generation using \texttt{torch.profiler} with
both CPU and CUDA activity tracking and \texttt{record\_shapes=True}, export the Chrome trace, and sum raw
kernel-execution durations directly from that trace. Appendix A explains why this direct summation is
necessary, rather than relying on the profiler's built-in aggregate API.

\textbf{(3) Scheduling simulation.} Using the batch service-time samples collected at batch size 8 in step
(2), run a 20,000-request Monte Carlo simulation. Draw Poisson inter-arrival times at each of five tested
utilization levels ($\rho \in \{0.30, 0.50, 0.70, 0.85, 0.95\}$). Resample service times from the measured
empirical distribution, rather than assuming an idealized distribution such as exponential, and record the
resulting latency percentiles. This step needs no GPU.

\textbf{(4) Correctness evaluation.} Run single-request inference over 100 GSM8K test problems at
gen\_length$=$512. Extract the final numeric answer from each generation, and compute exact-match accuracy
against the GSM8K reference answers.

\subsection{Evaluation and Expected Results}
\begin{itemize}
\item \textbf{Difficulty tiers:} 11 discrete denoising-step values ($178 + 29k$, $k = 0..10$) on GSM8K, with
step count predicting latency almost perfectly ($r > 0.99$).
\item \textbf{Variance under D2F:} GSM8K coefficient of variation in the range 0.25 to 0.29, measured under
a 512-token generation ceiling for the D2F-adapted model. The base model, without the adapter, shows a CV
roughly 2.5 to 3 times lower on identical prompts under the same ceiling.
\item \textbf{Admission-time prediction:} $R^2 < 0.15$ for every tested early signal against the eventual
step-count tier, meaning none of them are useful for admission-time scheduling.
\item \textbf{Generation-budget sensitivity:} Truncation rate falling from close to 100\% at
gen\_length$=$128 to a low single-digit percentage by gen\_length$=$512, with accuracy leveling off at the
same point.
\item \textbf{Batching throughput:} Synchronized batching scales throughput super-linearly relative to
per-request dispatch, whose overhead tracks the batch size closely.
\item \textbf{GPU profiling:} GPU kernel time makes up well under half of single-request wall-clock latency,
with host-side dispatch overhead dominating the rest.
\item \textbf{Scheduling behavior:} Mean latency under fixed-fill batching is U-shaped in utilization
$\rho$, with the minimum sitting near $\rho \approx 0.6$ to $0.8$.
\item \textbf{Correctness:} GSM8K exact-match accuracy in the 65 to 80\% range for LLaDA-8B with D2F, at
gen\_length $\geq$ 512, under single-request inference.
\end{itemize}
Exact numbers will vary somewhat from run to run, since the default decoding configuration uses a
temperature above 0. Results are expected to fall within the ranges above, rather than reproduce our exact
point estimates. Our own repeated measurements show this kind of run-to-run drift too, for example the
shift we observed when moving from an 8-batch to a 32-batch service-time sample in the scheduling
experiment, discussed in the Limitations section of the main paper.

\subsection{Experiment Customization}
The main settings a user can adjust are the generation-length budget, block size, the decode and
block-add confidence thresholds, batch size, and sampling temperature. Lowering the sample size for any
experiment trades statistical precision for shorter GPU time. We used sample sizes between 32 and 100
depending on the experiment, noted next to each table in the main paper, as a practical balance against
this submission's H200 time budget.

\subsection{Notes on Known Limitations}
Three limitations are worth restating here in concrete, reproduction-relevant terms.

First, the scheduling simulation was originally built on an 8-batch service-time sample. A later
replication using 32 batches shifted the estimated service capacity by about 24 percent, which tells us
the smaller sample had not yet converged. Anyone reproducing the scheduling results should collect a
larger service-time sample than the one used here before treating the percentile estimates as precise.

Second, we attempted a working slot-based batching design, one that mixes shared forward passes with
independent per-request completion, but did not get it to a fully correct state within this artifact. The
corresponding discussion of that design in the main paper is therefore analytical, not something we have
empirically validated.

Third, we could not complete direct empirical quality evaluation at batch sizes 8 and 16 within this
artifact's scope. The main paper instead gives a structural argument for why quality should hold at those
batch sizes, and this remains the clearest open item for future work.

This appendix is intentionally consistent with the Limitations section in the main paper. Every open item
mentioned there is restated here with the specific consequence it has for someone trying to reproduce or
extend this work.

\end{document}